\documentclass[journal]{IEEEtran}

\ifCLASSINFOpdf
\else
   \usepackage[dvips]{graphicx}
\fi
\usepackage{url}
\usepackage{graphicx}

\usepackage{algorithm}
\usepackage{algorithmic}

\usepackage{amsfonts, amsmath}

\usepackage{multirow}%
\usepackage{threeparttable}

\begin{document}

\title{UP-Diff: Latent Diffusion Model for Remote Sensing Urban Prediction}

\author{Zeyu Wang, \IEEEmembership{Student Member, IEEE}, Zecheng Hao, Jingyu Lin, Yuchao Feng, and Yufei Guo
\thanks{This work is supported by grants from the National Natural Science Foundation of China under contracts No.12202412 and No.12202413.}
\thanks{Zeyu Wang, Jingyu Lin, and Yufei Guo are with Intelligent Science \& Technology Academy of CASIC, China (e-mail: wangzeyu2020@zju.edu.cn, jingyu.lin@monash.edu, yfguo@pku.edu.cn). Zeyu Wang is also with College of Information Science \& Electronic Engineering, Zhejiang University, China.}
\thanks{Zecheng Hao is with School of Computer Science, Peking University, China (e-mail: haozecheng@pku.edu.cn).}
\thanks{Yuchao Feng is with College of Engineering, Westlake University, China (e-mail: fengyuchao@wioe.westlake.edu.cn).}
\thanks{\emph{Corresponding author: Yufei Guo}.}
}

\maketitle

\begin{abstract}
This study introduces a novel Remote Sensing (RS) Urban Prediction (UP) task focused on future urban planning, which aims to forecast urban layouts by utilizing information from existing urban layouts and planned change maps.
To address the proposed RS UP task, we propose UP-Diff, which leverages a Latent Diffusion Model (LDM) to capture position-aware embeddings of pre-change urban layouts and planned change maps.
In specific, the trainable cross-attention layers within UP-Diff's iterative diffusion modules enable the model to dynamically highlight crucial regions for targeted modifications.
By utilizing our UP-Diff, designers can effectively refine and adjust future urban city plans by making modifications to the change maps in a dynamic and adaptive manner.
Compared with conventional RS Change Detection (CD) methods, the proposed UP-Diff for the RS UP task avoids the requirement of paired pre-change and post-change images, which enhances the practical usage in city development.
Experimental results on LEVIR-CD and SYSU-CD datasets show UP-Diff's ability to accurately predict future urban layouts with high fidelity, demonstrating its potential for urban planning.
Code and model weights are available at https://github.com/zeyuwang-zju/UP-Diff.
\end{abstract}

\begin{IEEEkeywords}
   Remote Sensing (RS), Urban Prediction (UP), UP-Diff, Latent Diffusion Model (LDM), cross-attention.
\end{IEEEkeywords}

\IEEEpeerreviewmaketitle

\section{Introduction}

\IEEEPARstart{U}{rban} Prediction (UP) is a crucial field in city development, which forecasts and analyzes future trends in urban growth \cite{zhang2009data, triantakonstantis2012urban, aburas2016simulation}.
It helps researchers address the challenges arising from population growth \cite{bolorinos2020consumption}, transportation congestion \cite{williamson2002remote}, and climate changes \cite{pan2015haze, liu2016review}.
Currently, Remote Sensing (RS) technology plays a critical role in city development by capturing RS images about detailed urban landscapes \cite{schowengerdt2006remote, campbell2011introduction}, which are essential for monitoring the urban growth.
By leveraging the RS technology, researchers can significantly improve the accuracy and efficiency of urban layout prediction, enabling them to make well-informed decisions to optimize the urban development strategies.

\begin{figure}[!tbp]
   \centering
   \includegraphics[width=1\linewidth]{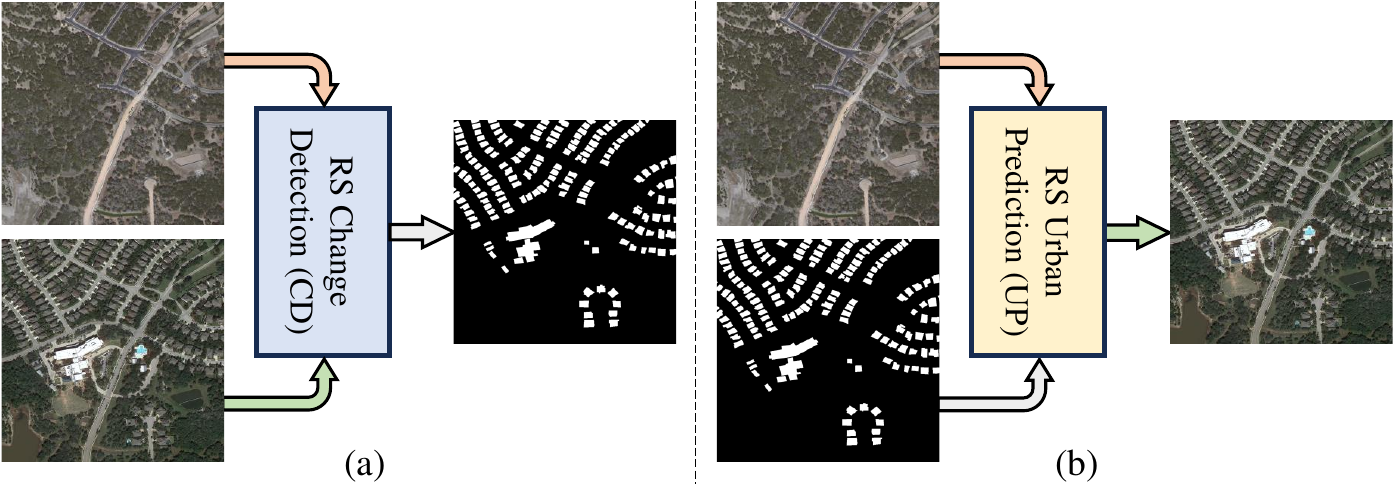}
   \caption{Illustration of the difference between (a) conventional RS Change Detection (CD) and (b) our proposed RS Urban Prediction (UP).}
   \label{intro}
\end{figure}

Despite the advancements in RS technology for city development, current works normally focus on the RS Change Detection (CD) task.
As illustrated in Fig. \ref{intro} (a), RS CD aims to identify the differences between two images captured at different time points of the same geographic area from the aerial view \cite{singh1989review, bai2023deep}.
However, in practical urban planning, our goal is to forecast future urban layouts based on existing urban layouts and proposed modifications \cite{liu2024diffusion}.
Current methods for RS CD primarily focus on detecting the layout changes rather than predicting future urban developments.
In addition, the requirement for paired pre-change and post-change images poses a challenge for RS data collection.
Collection of paired RS images must be guaranteed at long intervals, and the position and orientation of the probe must be accurate.
Therefore, it not only increases the cost of data acquisition and image alignment but also impedes effective urban planning.

In this paper, we propose a novel RS-based task focusing on urban planning, called RS Urban Prediction (UP).
As illustrated in Fig. \ref{intro} (b), the main objective of the proposed RS UP task is to forecast future urban layouts by leveraging information from existing urban layouts and planned change maps.
To address the challenge of the proposed task, we present UP-Diff, which is a novel Latent Diffusion Model (LDM) building upon the Stable Diffusion (SD) \cite{rombach2022high}.
Firstly, UP-Diff incorporates a trainable ConvNeXt model \cite{liu2022convnet} to encode the pre-change urban layouts and planned change maps into position-aware embeddings.
Subsequently, these embeddings are fed into the trainable cross-attention layers within the iterative diffusion modules of UP-Diff.
It enables the model to dynamically focus on the specific regions crucial for urban planning.
Meanwhile, we transfer the pre-trained SD model weights to mitigate the challenge of limited annotated RS image pairs.
Our main contributions are as follows:
\begin{itemize}
   \item We propose the RS Urban Prediction (UP) task to forecast future urban layouts based on existing urban RS layouts and planned change maps. To our knowledge, this is the first work on RS-based urban prediction.
   \item We also propose UP-Diff, which is a latent diffusion model with iterative layout-aware attention mechanism to dynamically enhance the critical urban regions.
   \item Extensive experiments on LEVIR-CD \cite{chen2020spatial} and SYSU-CD \cite{shi2021deeply} datasets demonstrate that UP-Diff can accurately predict the future urban layouts with high fidelity.
\end{itemize}

\section{Related Works}

\subsection{Remote Sensing Change Detection}
Current research on RS CD normally adopts deep learning-based models.
Convolutional Neural Networks (CNNs) have been widely applied on this task \cite{zhang2020deeply, chen2020dasnet, wang2021ads}.
Later, Transformer-based models have also shown the ability to capture long-range context and relationships between different positions \cite{chen2021remote, feng2023change}.
Recently, the advanced diffusion models, known for their effectiveness in capturing complex patterns in RS images, have also shown promise in RS CD \cite{bandara2022ddpm, wen2024gcd}.

\subsection{Diffusion Models}
Diffusion models have emerged as the leading choice for image generation, which originate from Denosing Diffusion Probabilistic Models (DDPMs) \cite{sohl2015deep, ho2020denoising}.
Building upon this foundation, the Latent Diffusion Model (LDM) \cite{rombach2022high} has pushed the boundaries by using a low-dimensional latent space \cite{esser2021taming}.
Current works on diffusion models normally guide the generation given text conditional information \cite{ramesh2022hierarchical, saharia2022photorealistic, rombach2022high}.
Moreover, recent advancements have been made in the field of layout-to-image diffusion models, where the spatial controls serve as the input conditions \cite{spatext, layoutdiffusion, li2023gligen, lin2024mirrordiffusion}.

\section{Methods}

\begin{figure}[!tbp]
   \centering
   \includegraphics[width=1\linewidth]{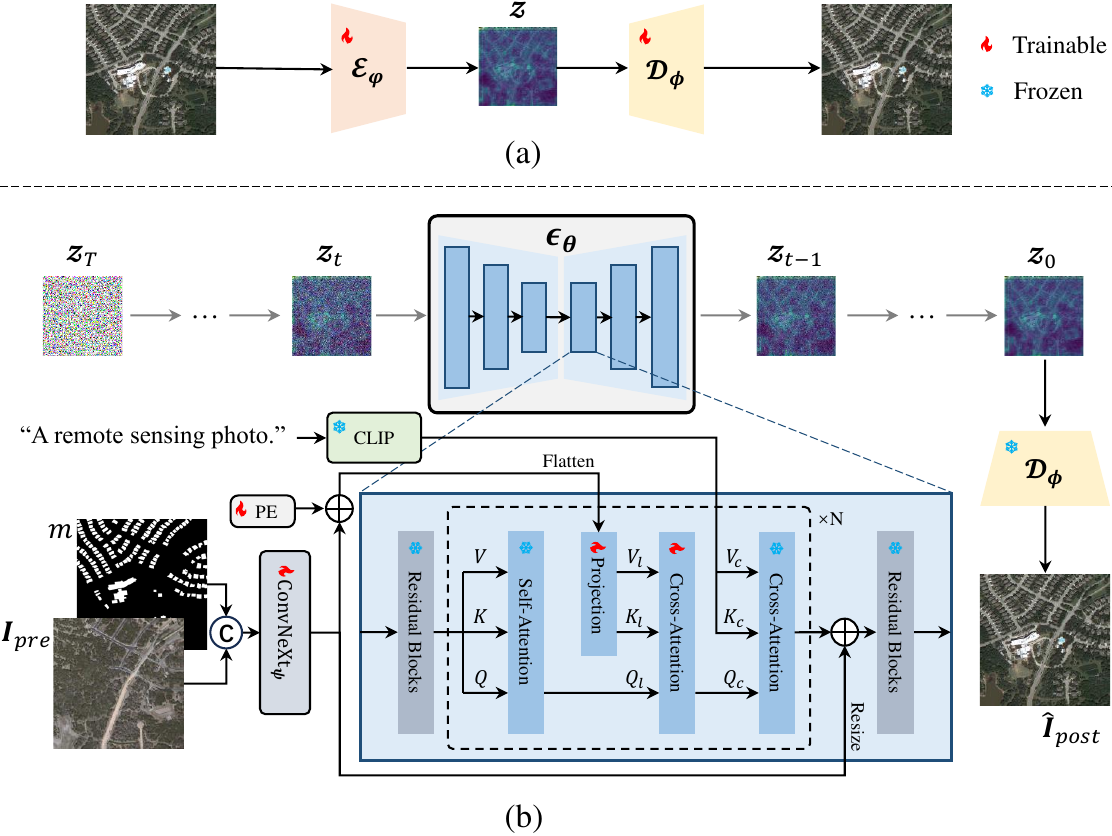}
   \caption{Illustration of our proposed UP-Diff for Remote Sensing Urban Prediction.
   (a) Training of the autoencoder for reconstruction.
   (b) Training of UP-UNet for latent diffusion and denoising. C denotes the concatenation.
   }
   \label{model}
\end{figure}

\subsection{Overview}
Our proposed UP-Diff is illustrated in Fig. \ref{model}.
To enhance the modeling efficiency, UP-Diff follows LDM \cite{rombach2022high}, which adopts an autoencoder to encode the input image into the latent space $\boldsymbol{z}$ and learn the representation through a reconstruction process.
The autoencoder comprises the encoder $\mathcal{E}_\varphi $ and the decoder $\mathcal{D}_\phi $.
In order to augment the model's comprehension of the RS layout and target alterations, we incorporate the embedded conditions through iterative attention-based blocks within the UP-UNet in the latent space.
It enables our UP-Diff model to dynamically prioritize specific target urban regions.
To our knowledge, UP-Diff is the first diffusion model for the proposed RS UP task, where the iterative diffusion process generates high-fidelity post-change RS images.

\begin{figure*}[!tbp]
   \centering
   \includegraphics[width=1\linewidth]{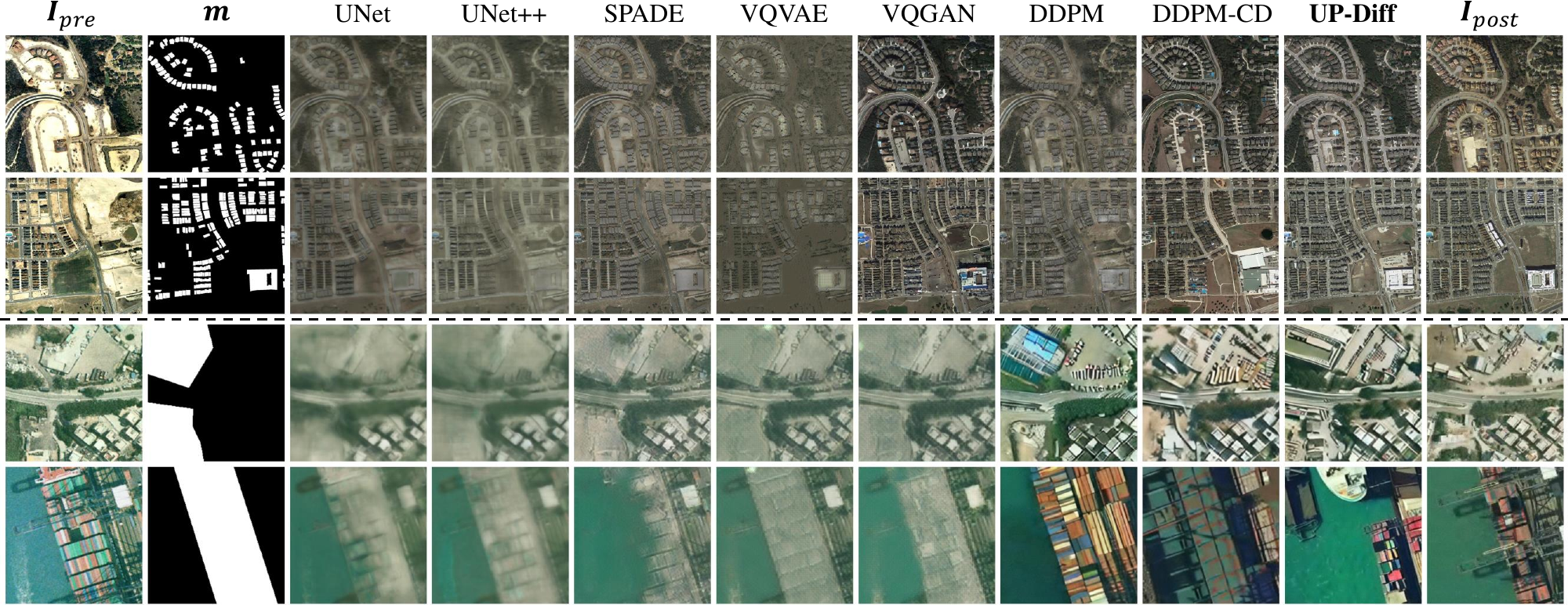}
   \caption{Qualitative results of the baseline methods and our UP-Diff on LEVIR-CD and SYSU-CD datasets for the proposed RS UP task.
   }
   \label{qualitative}
\end{figure*}

\subsection{Condition Embedding}
Condition embedding is critical in our proposed UP-Diff.
Firstly, the text condition $\boldsymbol{c}$ is obtained by embedding ``A remote sensing photo.'' via the CLIP text encoder \cite{radford2021learning}.
Utilizing the cross-attention layer from the SD model by incorporating $\boldsymbol{c}$ greatly enhances the awareness of RS features.

Meanwhile, the layout condition $\boldsymbol{l}$ is obtained by embedding the concatenated pre-change image $\boldsymbol{I}_{pre}$ and change map $\boldsymbol{m}$ by a trainable ConvNeXt model \cite{liu2022convnet} with parameters $\boldsymbol{\psi}$:
\begin{equation}
   \boldsymbol{l} = \text{ConvNeXt}_{\boldsymbol{\psi}} (\text{Concat}[\boldsymbol{I}_{pre}, \boldsymbol{m}]),
\end{equation}
where the semantic layout feature $\boldsymbol{l}$ is added after the iterative attention-based blocks.
The ConvNeXt model has been pre-trained on large-scale natural image datasets.
In addition, the tokenized layout condition $\boldsymbol{l}^{*}$ is derived by flattening $\boldsymbol{l}$:
\begin{equation}
   \boldsymbol{l}^{*} = \text{Flatten} (\boldsymbol{l}) + \text{PE},
\end{equation}
where PE denotes a trainable position embedding to enhance the position-aware information.
The tokenized $\boldsymbol{l}^{*}$ serves as a suitable form for the gated cross-attention mechanism.

\subsection{UP-UNet}
Our proposed UP-UNet, denoted as $\boldsymbol{\epsilon}_{\boldsymbol{\theta}}$, serves as the core component for the latent denoising iterations.
UP-UNet inherits the spatial structure of UNet \cite{ronneberger2015u}, where the basic module consists of the residual convolutional blocks \cite{he2016deep} and iterative attention-based layers \cite{vaswani2017attention}.
We keep the original layers from SD model \cite{rombach2022high} and frozen the model weights to fully utilize the generative model pre-trained on large-scale natural images.
As illustrated in Fig. \ref{model}, the UP-UNet is mainly composed of cascaded self-attention and cross-attention layers, which can be formulated as below:
\begin{equation}
   \text{Attn}(Q, K, V) = \text{softmax}\left(\frac{Q\cdot K^T}{\sqrt{d_k}}\right)\cdot V,
\end{equation}
where the matrices $Q$, $K$, and $V$ are derived from linear projection of the input features.
The token dimension of the matrix $K$ is denoted by $d_k$.
In self-attention mechanism, the matrices $Q$, $K$, and $V$ are derived from the same input feature.
In cross-attention mechanism, $Q$ is derived from one modality, while $K$ and $V$ are obtained from another modality.
The cross-attention layer is followed by the feed-forward layer to increase the representation ability, which is omitted in Fig. \ref{model}.

In specific, the trainable cross-attention layer in the middle of UP-UNet block incorporates a gated mechanism, which focuses on the target RS layout regions.
The gated cross-attention layer can be formulated as:
\begin{equation}
   Q_c = Q_l + \lambda \cdot \text{tanh}(\gamma)(\text{Attn}(Q_l, K_l, V_l)),
\end{equation}
where $\lambda$ is a hyperparameter and $\gamma$ is a learnable parameter initialized as 0.
The tokens $K_l$ and $V_l$ are obtained by the linear projection from $\boldsymbol{l}^{*}$.
Inside the iterative diffusion blocks of UP-UNet, a trainable cross-attention layer is introduced between the fixed self-attention and cross-attention layers, enabling the model to anticipate the layout of change area.
Through the implementation of a gated mechanism, the model can dynamically adjust the attention weights to enhance its perception on the target urban layout.

\subsection{Training \& Sampling Formulation}
We first fine-tune the autoencoder for reconstruction on the RS image datasets.
During the diffusion training process, the model weights of the autoencoder are freezed, and the latent $\boldsymbol{z}_0$ is obtained by encoding the post-change image $\boldsymbol{I}_{post}$ through $\mathcal{E}_\varphi $.
After that, the noise-contaminated latent features $\boldsymbol{z}_1, \boldsymbol{z}_2, ..., \boldsymbol{z}_T$ are achieved via a Markov chain:
\begin{equation}
   q\left(\boldsymbol{z}_t \mid \boldsymbol{z}_{t-1}\right)=\mathcal{N}\left(\boldsymbol{z}_t ; \sqrt{1-\beta_t} \boldsymbol{z}_{t-1}, \beta_t \mathbf{I}\right),
\end{equation}
where the hyperparameter $\beta_t$ corresponds to the variance of the Gaussian distribution, and it undergoes a linear increase as $t$.
Meanwhile, we can directly obtain $\boldsymbol{z}_t$ based on $\boldsymbol{z}_0$ and $\beta_t$ via the reparameterization \cite{kingma2013auto}:
\begin{equation}
   \boldsymbol{z}_t = \sqrt{\overline{\alpha}_t }\boldsymbol{z}_0 + \sqrt{1-\overline{\alpha}_t } \boldsymbol{\epsilon},
\end{equation}
where $\boldsymbol{\epsilon} \sim \mathcal{N}(\mathbf{0}, \mathbf{I})$ denotes the randomly sampled Gaussian noise.
Meanwhile, $\alpha_t = 1 - \beta_t$ and $\overline{\alpha}_t = \prod_{i=1}^t \alpha_i$.
Therefore, $\boldsymbol{z}_T$ converges to standard normal distribution $\mathcal{N}(\mathbf{0}, \mathbf{I})$ when $T \rightarrow \infty$.
The training objective of our proposed UP-UNet, denoted as $\boldsymbol{\epsilon}_{\boldsymbol{\theta}}$, can be represented as:
\begin{equation}
   \min _{\boldsymbol{\theta}} \mathcal{L}_{\text{UP-Diff}}=\mathbb{E}_{\boldsymbol{z}_t, \boldsymbol{\epsilon} \sim \mathcal{N}(\mathbf{0}, \mathbf{I}), t, \boldsymbol{c}}\left[\left\|\boldsymbol{\epsilon}-\boldsymbol{\epsilon}_{\boldsymbol{\theta}}\left(\boldsymbol{z}_t, t, \boldsymbol{l}, \boldsymbol{c}\right)\right\|_2^2\right],
\end{equation}
where $t \sim \text{Uniform}(\{1, ..., T\})$ is a random time step.
$\boldsymbol{l}$ and $\boldsymbol{c}$ denote the layout condition and text condition, respectively.

During the inference process, we can reverse the diffusion process into the denoising process.
Starting from a Gaussian noise $\boldsymbol{z}_T$, our proposed UP-UNet gradually generates less noisy $\boldsymbol{z}_{T-1}, \boldsymbol{z}_{T-2}, ..., \boldsymbol{z}_{0}$ by the following iteration:
\begin{equation}
   \boldsymbol{z}_{t-1}=\frac{1}{\sqrt{\alpha_t}}\left(\boldsymbol{z}_t-\frac{1-\alpha_t}{\sqrt{1-\bar{\alpha}_t}} \boldsymbol{\epsilon}_\theta\left(\boldsymbol{z}_t, t, \boldsymbol{l}, \boldsymbol{c} \right)\right)+\beta_t \boldsymbol{\epsilon}.
\end{equation}

Finally, we obtain the denoised $\boldsymbol{z}_{0}$, which is then decoded by $\mathcal{D}_\phi $ into the generated post-change image $\widehat{\boldsymbol{I}} _{post}$.

\section{Experiments}

\subsection{Implementations}
We implement UP-Diff using PyTorch 1.13.1 on Intel(R) Xeon(R) Platinum 8255C CPU@2.50GHz and two NVIDIA Tesla V100 GPUs with CUDA 12.2.
AdamW optimizer is adopted with the learning rate of 5e-5 and 10,000 warmup training iterations.
The batch sizes per GPU are set to 2 and 4 for training the autoencoder and the UP-Diff, respectively.
Additionally, we employ random flips and crops on the RS images for data augmentation.

\subsection{Datasets}
We conduct experiments on two datasets with different RS scenes and resolutions to demonstrate the generalizability:
\begin{itemize}
   \item LEarning, VIsion, and Remote sensing Dataset (LEVIR-CD) \cite{chen2020spatial} contains 445, 64, and 128 RS image pairs for training, validation, and testing, respectively, captured in Texas, USA. We conduct the experiments on the dataset with the large resolution of 512 $\times$ 512.
   \item Sun Yat-Sen University Dataset (SYSU-CD) \cite{shi2021deeply} is a large-scale RS image dataset gathered in Hong Kong, China, containing 12,000/4,000/4,000 samples for training/validation/testing, respectively. We adopt the official resolution of 256 $\times$ 256 for experiments.
\end{itemize}

\subsection{Baselines}
We compare our UP-Diff with the following baselines listed in Table \ref{lpips}.
For the conventional deep learning models, we concatenate $\boldsymbol{I}_{pre}$ and $\boldsymbol{m}$ as input for end-to-end training.
For the diffusion-based models, we implement them by replacing the original $\boldsymbol{I}_{post}$ with $\boldsymbol{m}$ as the input conditions.

\subsection{Qualitative Comparison}

The qualitative results of the baseline methods and our UP-Diff on LEVIR-CD and SYSU-CD datasets are shown in Fig. \ref{qualitative}.
It can be seen that pre-change images $\boldsymbol{I}_{pre}$ (the first column) are typically captured on barren land, while the post-change images $\boldsymbol{I}_{post}$ (the last column) display neatly arranged buildings.
The baseline models employing end-to-end training, including UNet, UNet++, SPADE, and VQVAE, can generate the post-change images with general building outlines but lack fidelity and diversity.
The advanced VQGAN, DDPM, and DDPM-CD models produce high-fidelity images, but some texture details do not accurately align with the actual transformations depicted in the change maps.
In contrast, our proposed UP-Diff model can produce high-quality post-change images that faithfully capture target urban modifications, such as the specific buildings, main roads, and vessels on the water.

\subsection{Quantitative Comparison}

\begin{table}[!tbp]
   \small
   \setlength\tabcolsep{8.0pt}
   \renewcommand\arraystretch{1.0}
   \caption{Quantitative comparison on the generated image quality.}
   \centering
   \label{lpips}
   \begin{threeparttable}
   \begin{tabular}{c|cc|cc}
     \hline
      \multirow{2}{*}{Model} & \multicolumn{2}{c|}{LEVIR-CD} & \multicolumn{2}{c}{SYSU-CD} \\
      \cline{2-5}
      & LPIPS$\downarrow$ & FID$\downarrow$ & LPIPS$\downarrow$ & FID$\downarrow$ \\
     \hline
     \hline
      UNet \cite{ronneberger2015u} & 0.504 & 196.28 & 0.521 & 139.59    \\
      UNet++ \cite{zhou2018unet++} & 0.497 & 185.48 & 0.523 & 120.05    \\
      SPADE \cite{park2019semantic}& 0.423 & 142.58 & 0.436 & 66.28    \\
      VQVAE \cite{van2017neural}   & 0.406 & 185.45 & 0.477 & 73.20    \\
      VQGAN \cite{esser2021taming} & 0.402 & 179.22 & 0.478 & 60.56    \\
      DDPM \cite{ho2020denoising}  & 0.420 & 134.63 & 0.430 & 61.03    \\
      DDPM-CD \cite{bandara2022ddpm} & 0.418 & 144.38 & 0.414 & 49.80  \\
      \textbf{UP-Diff} & \textbf{0.342} & \textbf{117.79} & \textbf{0.400} & \textbf{34.57} \\
     \hline
   \end{tabular}
   \begin{tablenotes}
      \footnotesize
      \item $\downarrow$ denotes that lower values are better. 
   \end{tablenotes}
   \end{threeparttable}
\end{table}

We evaluate the models from two quantitative aspects.
On one hand, we evaluate the quality of the generated images using Learned Perceptual Image Patch Similarity (LPIPS) \cite{zhang2018unreasonable} and Fréchet Inception Distance (FID) \cite{heusel2017gans}.
On the other hand, inspired by the assessment in semantic image synthesis \cite{park2019semantic}, we utilize a pre-trained RS CD model (DMINet \cite{feng2023change}) to test the RS CD metrics based on original input $\boldsymbol{I}_{pre}$ and output $\widehat{\boldsymbol{I}} _{post}$.
The metrics include Precision (Pre.), Recall (Rec.), F1-score (F1), and Intersection over Union (IoU). 

\begin{figure}[!tbp]
   \centering
   \includegraphics[width=1\linewidth]{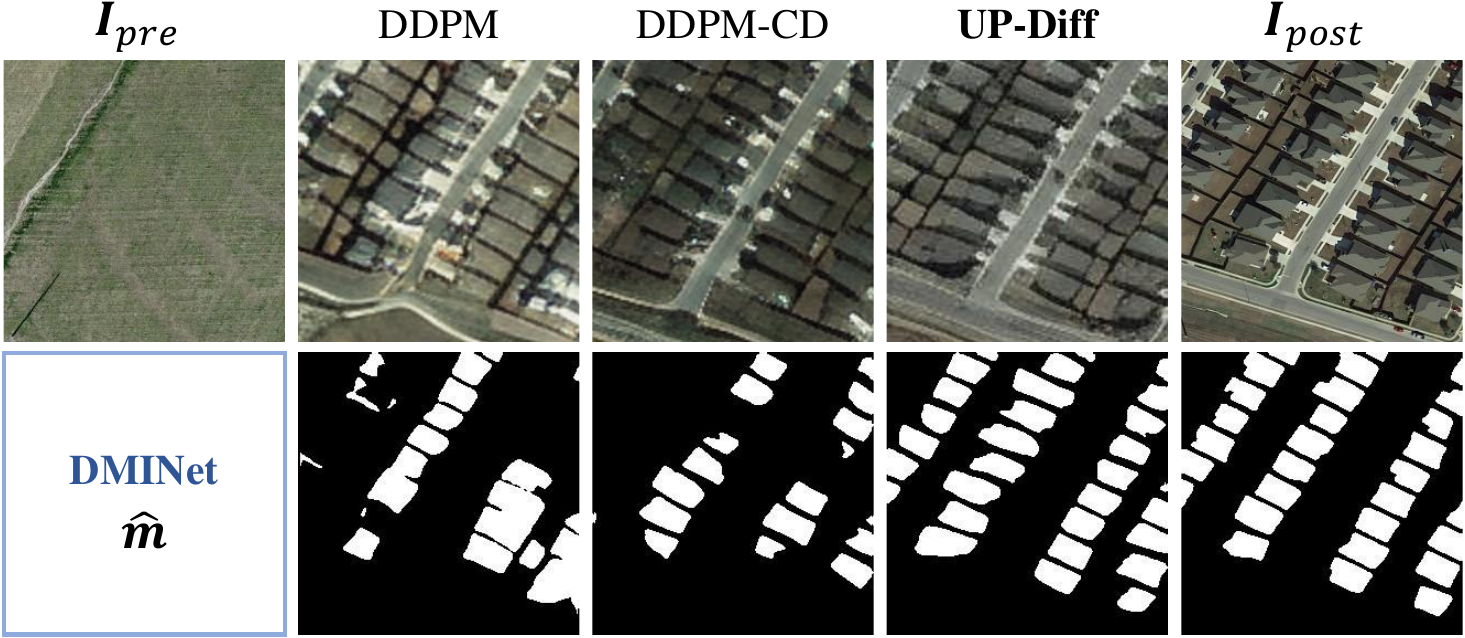}
   \caption{Qualitative comparison on the generated images for RS CD task.
   }
   \label{dminet}
\end{figure}

The results in Table \ref{lpips} demonstrate that our UP-Diff model achieves the best LPIPS and FID scores across both datasets, with scores of 0.342 LPIPS and 117.79 FID on the LEVIR-CD dataset, and 0.400 LPIPS and 34.57 FID on the SYSU-CD dataset, respectively.
The diffusion-based model DDPM-CD also exhibits commendable performance, achieving LPIPS scores of 0.418 and 0.414 on the two datasets, respectively.
Meanwhile, from Table \ref{quantitative1}, we can see that when we use the generated post-change images $\widehat{\boldsymbol{I}} _{post}$ for evaluation on the RS CD task, the evaluation scores all demonstrate our excellent performance.
For example, UP-Diff achieves the Pre. score of 92.23\% and F1 score of 82.41\% on LEVIR-CD dataset, respectively. 
However, these baseline methods exhibit obvious inferior performance when contrasted with our method.
We also show an example of the generated change maps through DDPM, DDPM-CD, and our UP-Diff in Fig. \ref{dminet}.
It can be seen that the result of our UP-Diff aligns well with the ground-truth $\boldsymbol{I}_{post}$, while the results from baseline methods exhibit differing degrees of deviation from $\boldsymbol{I}_{post}$.

\begin{table}[!tbp]
   \small
   \setlength\tabcolsep{9.0pt}
   \renewcommand\arraystretch{1.0}
   \begin{threeparttable}
   \caption{Quantitative comparison on the generated image for RS CD task. The results are recorded in (\%).}
   \centering
   \label{quantitative1}
   \begin{tabular}{c|c|cccc}
     \hline
     & Model & Pre.$\uparrow$ & Rec.$\uparrow$ & F1$\uparrow$ & IoU$\uparrow$ \\
     \hline
     \hline
     \multirow{8}{*}{\rotatebox{90}{LEVIR-CD}} 
     & UNet \cite{ronneberger2015u} & 55.54 & 50.41 & 49.70 & 47.90   \\
     & UNet++ \cite{zhou2018unet++} & 64.39 & 55.79 & 59.19 & 53.51   \\
     & SPADE \cite{park2019semantic}& 80.33 & 64.52 & 71.46 & 62.32   \\
     & VQVAE \cite{van2017neural}   & 70.00 & 59.82 & 64.96 & 57.43   \\
     & VQGAN \cite{esser2021taming} & 84.56 & 69.63 & 74.79 & 65.43  \\
     & DDPM \cite{ho2020denoising}  & 74.58 & 68.13 & 77.23 & 69.74   \\
     & DDPM-CD \cite{bandara2022ddpm} & 86.54 & 72.15 & 77.37 & 67.93  \\
     & \textbf{UP-Diff} & \textbf{92.23} & \textbf{76.55} & \textbf{82.41} & \textbf{73.35} \\
     \hline
     \hline    
     \multirow{8}{*}{\rotatebox{90}{SYSU-CD}} 
     & UNet \cite{ronneberger2015u} & 77.23 & 84.63 & 79.10 & 66.26    \\
     & UNet++ \cite{zhou2018unet++} & 77.71 & 85.29 & 79.62 & 66.94    \\
     & SPADE \cite{park2019semantic}& 79.58 & 84.52 & 80.33 & 68.71   \\
     & VQVAE \cite{van2017neural}   & 77.43 & 84.97 & 78.62 & 66.89   \\
     & VQGAN \cite{esser2021taming} & 81.02 & 85.26 & 80.05 & 67.48   \\
     & DDPM \cite{ho2020denoising}  & 80.14 & 83.29 & 81.48 & 69.74   \\
     & DDPM-CD \cite{bandara2022ddpm} & 84.11 & 85.43 & 84.74 & 74.33  \\
     & \textbf{UP-Diff} & \textbf{86.82} & \textbf{86.10} & \textbf{86.45} & \textbf{76.86} \\
     \hline
   \end{tabular}
   \begin{tablenotes}
      \footnotesize
      \item $\uparrow$ denotes that higher values are better. 
   \end{tablenotes}
   \end{threeparttable}
\end{table}

\section{Conclusion}
In this work, we introduce a novel task called Remote Sensing (RS) Urban Prediction (UP) for forecasting future urban layouts.
Meanwhile, we propose the latent UP-Diff model tailored for the RS UP task, utilizing iterative layout-based gated cross-attention layers to focus on critical regions.
Our qualitative and quantitative results on two datasets illustrate the superiority of UP-Diff over baseline methods in the RS UP task, including conventional deep learning and diffusion-based approaches.
The post-change images generated by UP-Diff align well with ground-truth images during migration to the RS Change Detection (CD) task with a pre-trained model.
In the future, we aim to develop more efficient methods with mobility for the proposed RS UP task.
Furthermore, we plan to introduce a novel pluralistic RS UP task, where the generated post-change images can exhibit great diversity, contributing to flexible urban planning initiatives.

\bibliographystyle{IEEEtrans}
\bibliography{citation.bib}

\end{document}